\documentclass{article}

\usepackage[preprint, nonatbib]{neurips_2022}

\usepackage[utf8]{inputenc} % allow utf-8 input
\usepackage[T1]{fontenc}    % use 8-bit T1 fonts
\usepackage{hyperref}       % hyperlinks
\usepackage{url}            % simple URL typesetting
\usepackage{booktabs}       % professional-quality tables
\usepackage{amsfonts}       % blackboard math symbols
\usepackage{nicefrac}       % compact symbols for 1/2, etc.
\usepackage{microtype}      % microtypography
\usepackage{xcolor}         % colors
\usepackage{amssymb}
\usepackage{multirow}
\usepackage{array}
\usepackage{xcolor}
\usepackage{caption}
\usepackage{subcaption}
\usepackage{amsmath}
\usepackage{booktabs}
\usepackage{pifont}
\usepackage{array, makecell} %
\usepackage{diagbox}
\usepackage{verbatim}
\usepackage{hyperref}
\usepackage{graphicx}
\newcommand{\ppm}{\,\scriptsize$\pm$}

\def\gX{{\mathcal{X}}}
\def\gS{{\mathcal{S}}}
\def\gT{{\mathcal{T}}}
\def\gL{{\mathcal{L}}}
\def\gA{{\mathcal{A}}}
\def\gB{{\mathcal{B}}}
\def\gC{{\mathcal{C}}}
\def\gD{{\mathcal{D}}}
\def\gF{{\mathcal{F}}}
\def\gU{{\mathcal{U}}}

\def\vx{{\mathbf{x}}}
\def\vu{{\mathbf{u}}}
\def\vz{{\mathbf{z}}}
\def\vy{{\mathbf{y}}}

\def\sR{{\mathbb{R}}}

\newcommand{\mypar}[1]{\vspace{0.25em}\noindent\textbf{#1}~}
\newcolumntype{C}[1]{>{\centering\let\newline\\\arraybackslash\hspace{0pt}}m{#1}}

\newcommand{\xmark}{\ding{55}}
\newcommand{\cmark}{\ding{51}}
\newcommand{\argmin}{\operatornamewithlimits{argmin}}

\title{Harmonizing Flows: Unsupervised MR harmonization based on normalizing flows}

\author{%
Farzad Beizaee$^{1,2*}$ \quad Christian Desrosiers$^{1}$ \quad Gregory A. Lodygensky$^{2,3}$ \quad Jose Dolz$^1$ \\
$^1$\text{ÉTS Montreal} \\$^2$CHU Sainte-Justine, University of Montreal \\\quad $^3$Canadian Neonatal Brain Platform, Montreal\\
$^*$\texttt{farzad.beizaee.1@ens.etsmtl.ca}
}

\begin{document}

\maketitle

\begin{abstract}
In this paper, we propose an unsupervised framework based on normalizing flows that harmonizes MR images to mimic the distribution of the source domain. The proposed framework consists of three steps. First, a shallow harmonizer network is trained to recover images  of the source domain from their augmented versions. A normalizing flow network is then trained to learn the distribution of the source domain. Finally, at test time, a harmonizer network is modified so that the output images match the source domain's distribution learned by the normalizing flow model. Our unsupervised, source-free and task-independent approach is evaluated on cross-domain brain MRI segmentation using data from four different sites. Results demonstrate its superior performance compared to existing methods. The code is available at \href{https://github.com/farzad-bz/Harmonizing-Flows}{https://github.com/farzad-bz/Harmonizing-Flows} 
\end{abstract}

\section{Introduction}
% \subsection{A Subsection Sample}
% Please note that the first paragraph of a section or subsection is
% not indented. The first paragraph that follows a table, figure,
% equation etc. does not need an indent, either.

%Understanding and interpreting the content of medical imaging data is of pivotal importance for the diagnostic, treatment planning and follow-up of many diseases. 
Deep learning models have become the \textit{de facto} solution for most image-based problems, including those in the medical domain. Despite significant progress, these models still suffer under distributional drift, and their performance largely degrades when they are applied to data obtained in different conditions. 

Clinical studies using magnetic resonance imaging (MRI) often have to deal with such large domain shifts. Due to the qualitative nature of the MRI acquisition process, generated images are sensitive to imaging devices, acquisition protocols, scanner artifacts, as well as to patient populations \cite{takao2011effect}. For instance, images from the same modality (e.g., T1-w) acquired from two different scanners with separate configurations will likely present noticeable differences, which can be considered a domain shift. Consequently, collecting a multi-center MRI dataset to address a particular clinical question does not guarantee a greater statistical power, as the increase in variance comes from a non-clinical source. Furthermore, this data heterogeneity can also hamper the generalizability of deep learning models, preventing their large dissemination. In particular, when trained on a specific site, such models are typically unable to provide similar performance for other centers.
%Deep learning is at the core of 

To alleviate this issue, image harmonization addresses the distributional shift problem from an image-to-image mapping perspective, where the objective is to transfer image contrasts across different domains. Nevertheless, most harmonization methods in the literature make strong assumptions that might hamper their scalability and usability in real-life scenarios. First, some methods must have access to source images during the adaptation, which may no longer be available. Labels associated with the downstream task may also be required in other approaches. Finally, most harmonization techniques need to know the target domains during training, while these domains are often unknown. %In this work, we seek to develop a harmonization method which is \textit{source-free} ($\gS\gF$), \textit{task-agnostic} ($\gT\gA$) and can handle \textit{unknown-domains} ($\gU\gD$). 

%%% CONTRIBUTIONS AS A BULLET POINTS %%%%

In this work, we make the following contributions:
\begin{itemize}
    \item We relax all these assumptions and present a novel MR harmonization method that is \textit{source-free} ($\gS\gF$), \textit{task-agnostic} ($\gT\gA$) and can handle \textit{unknown-domains} ($\gU\gD$) without requiring to be retrained for each target distribution. Indeed, our method only needs one domain and modality at training time, as opposed to existing approaches. 
    \item In particular, we propose to use a novel family of generative models, i.e., normalizing flows, which have shown to be a powerful method to model data distributions in generative tasks. We stress that leveraging normalizing flows to guide the adaptation of a harmonizer network has not been explored. %yet.
    \item In addition to the methodological novelty, our empirical results demonstrate that the our approach brings substantial improvements compared to existing techniques, while alleviating their weaknesses. 
    \item Furthermore, due to its task-agnostic nature and its capability to work under the \textit{unknown-domains} scenario, the proposed method can also be employed in the task of test-time adaptation. In this setting, our method largely outperforms a popular task-agnostic test-time adaptation strategy.
\end{itemize}

\section{Related work}

 \mypar{Image harmonization.} Several techniques have been proposed for the harmonization of  images in the medical domain, and particularly for MRI data. Classical post-processing steps, such as intensity histogram matching \cite{nyul2000new,shinohara2014statistical}, reduce the influence of biases across scanners, but may also remove informative local variations in intensity. Statistical approaches can model image intensity and dataset bias at the voxel level \cite{fortin2016removing,fortin2017harmonization,beer2020longitudinal}, however they must often be adjusted each time images from new sites are provided. Modern strategies for image harmonization, which are based on deep learning models, have shown to be a promising alternative for this problem \cite{dewey2019deepharmony,zhu2017unpaired,liu2021style,zuo2021information,delisle2021realistic,dinsdale2021deep}. Nevertheless, they make unrealistic assumptions that hamper the scalability of existing approaches to large scale multi-site harmonization tasks. First, images of the same target anatomy across multiple sites, commonly referred to as \textit{traveling subjects} are employed to identify intensity transformations between different sites \cite{dewey2019deepharmony}. This involves that a given number of subjects are scanned at every site or scanner required for training, a condition rarely met in practice. Second, another group of methods is limited to two domains \cite{zhu2017unpaired} and requires target domains to be known at training time \cite{zhu2017unpaired,liu2021style}. In addition, each time a new domain is added, these approaches must be fine-tuned in order to accommodate the characteristics of each domain. Calamity \cite{liu2021style} further needs paired multi-modal MR sequences, limiting even more its applicability to single modality scenarios. Last, task-dependent approaches leverage labels associated to each image for a given down-stream task \cite{delisle2021realistic,dinsdale2021deep}, thus optimizing the harmonization for this specific problem. Nevertheless, having access to large labeled datasets might be impractical due to the underlying labeling cost.

\mypar{Test-time Adaptation.} Our method also relates to the problem of test-time domain adaptation (TTA)~\cite{wang2020tent,sun2020test,boudiaf2022parameter} which aims to quickly adapt a pre-trained deep network to domain shifts during inference on test examples. One key difference between TTA and the well-known unsupervised domain adaption (UDA) problem is that, in TTA, the source examples are no longer available. One of the earliest TTA approaches, called TENT \cite{wang2020tent}, updates the affine transformation parameters of normalization layers by minimizing the Shannon entropy of predictions for test examples. In \cite{mummadi2021test}, this strategy is improved by optimizing a log-likelihood ratio instead of entropy, as well as by considering the normalization statistics of the test batch. The method named SHOT \cite{liang2020we} fine-tunes the entire feature extractor with a mutual information loss and uses pseudo-labels to provide additional test-time guidance. Instead of updating the network parameters, LAME \cite{boudiaf2022parameter} uses Laplacian regularization to do a post-hoc adaptation of the softmax predictions.

\mypar{Normalizing flows.}
Recently, normalizing Flows (NFs) have emerged as a popular approach for constructing probabilistic and generative models with tractable distributions~\cite{kobyzev2020normalizing}. NFs aim at transforming unknown complex distributions into simpler ones, for instance, a standard normal distribution. This is achieved by applying a sequence of invertible and differentiable transformations. While most existing literature has leveraged NFs for generative tasks 
(e.g., image generation \cite{ho2019flow++,kingma2018glow}, noise modeling \cite{abdelhamed2019noise}, graph modeling \cite{zang2020moflow}) and anomaly detection \cite{gudovskiy2022cflow,kirichenko2020normalizing}, recent evidence also suggests their usefulness for aligning a given set of source domains~\cite{grover2020alignflow,usman2020log}. %In contrast, our approach learns information from source domain data so that new unseen domains can be processed in the same way. 
To our knowledge, a single work has investigated NFs in the context of harmonization~\cite{wang2021harmonization}. However, it aimed at performing causal inference on pre-extracted features (brain ROI volume measures), and not image harmonization as in our work. Moreover, since extracting ROIs requires pixel-wise labels, the method in \cite{wang2021harmonization} is not task-agnostic. 

\section{Methodology}

\begin{figure}[t]
\centering
\includegraphics[width=\linewidth]{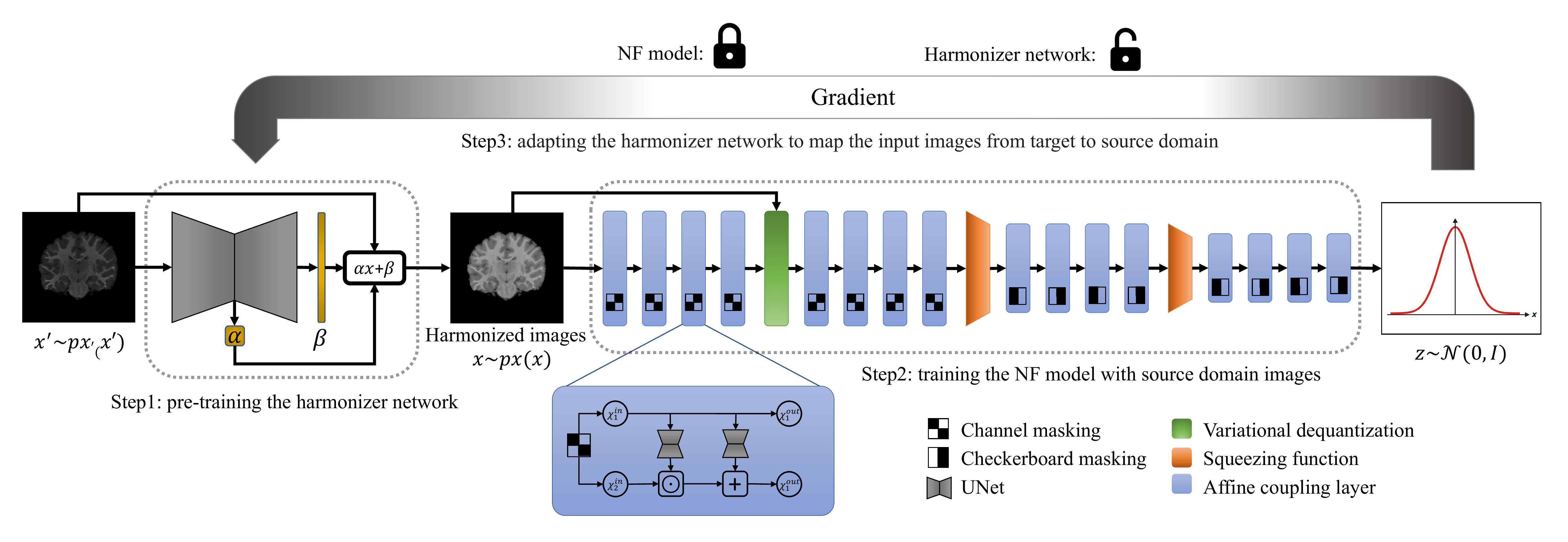}
     \caption{\textbf{Pipeline of the proposed Harmonizing Flows method.} Our approach consists of two steps. First, we employ normalizing flows (NFs) to capture the distribution of the source domain. During the second stage, the trained NFs are leveraged to update the parameters of a harmonizer network, which are updated in order to maximize the similarity between the harmonized outputs and the distribution learned by the NF. Note that steps 1 and 2 are not dependent on each other, and can therefore be performed in any order. %First, a NF-based model is trained to learn the source distribution (Section \ref{ssec:NFs}), while we also pre-train harmonizer  network (Section \ref{ssec:harmo}, Harmonizer).
     }
\label{fig:normalizing flow architecture.}
\end{figure}

We first define the problem addressed in our work. Let $\gX_\gS=\{\vx_{n}\}_{n=1}^{N}$ be a set of unlabeled images in the source domain $\gS$, where a given image $i$ is represented by $\vx_i \in \sR^{|\mathrm{\Omega}|}$ and $\mathrm{\Omega}$ denotes its spatial domain (i.e., $W\!\times\!H$). Similarly, we denote as $\gX_\gT=\{\vx_{n}\}_{n=1}^{M}$ the set of unlabeled images in a potential target domain $\gT$\footnote{Note that for simplicity, we assume here that there exists only a single domain. Nevertheless, our formulation is directly applied to $T$ different domains.}. The goal of unsupervised data harmonization is to find a mapping function $f_{\theta} : \gS \!\rightarrow\!\gT$ without having access to labeled images for any of the domains. In what follows, we present our NF-based solution for this problem, whose framework is depicted in Figure \ref{fig:normalizing flow architecture.}. %Our approach consists of two steps. First, we employ normalizing flows to capture the distribution of the source domain. During the second stage, the trained normalizing flows are leveraged to update the parameters of a harmonizer network, which are updated in order to maximize the similarity between the harmonized outputs and the distribution learned by the normalizing flow.

%Logical steps toward source-free and unsupervised harmonization are learning the distribution of the source domain and then changing the distribution of the target MRIs to follow the learned source distribution. We stick to the same procedure in this work using a three-step framework. First, NFs, which have shown an excellent capability of learning the exact likelihood of data, are leveraged to capture the distribution of the source domain. Second, an auto-encoder is considered as a harmonizer network and is pre-trained by reconstructing original images of the source domain from augmented images. Finally, using the images from the unseen target domain at the test time, the parameters of the harmonizer network are updated, so the harmonized outputs follow the distribution learned by the NF. In the following, each step is detailed.

\setcounter{footnote}{0} 

\subsection{Learning the source domain distribution}
\label{ssec:NFs}
We leverage Normalizing Flows (NFs) \cite{dinh2016density} to model the distribution of the source domain. NFs are a recent family of generative methods that can model a complex probability density $p_{x}(\vx)$ (i.e., the source) as a series of transformation functions, denoted as $g_{\phi}=g_1 \circ g_2 \circ \dots g_T$, applied on simpler and tractable probability density $p_{u}(\vu)$ (e.g., a standard multi-variate Gaussian distribution). We can express a source image as $\vx=g_{\phi}(\vu)$, where $\vu \sim p_{u}(\vu)$ and $p_{u}(\vu)$ is the base distribution of the flow model. An important requirement of the transformation function $g_{\phi}$ is that it must be \textit{invertible}, and both $g_{\phi}$ and $g_{\phi}^{-1}$ should be \textit{differentiable}. Under these conditions, the density of the original variable $\vx$ is well-defined and its likelihood can be computed exactly using the change of variables rule as:
\begin{equation}
\begin{aligned}
 \log  p_{\mathrm{x}}(\vx) \, &= \, \log p_{\mathrm{z}}\left(g^{-1}_{\phi}(\vx)\right) + \log\left|\operatorname{det} \big(\mathbf{J}_{g^{-1}_{\phi}}(\vx)\big)\right| \\
 &= \log p_{\mathrm{z}}\left(g^{-1}_{\phi}(\vx)\right) + \sum_{t=1}^T \log\left|\operatorname{det} \big(\mathbf{J}_{g^{-1}_{t}}(\vu_{t-1})\big)\right|
\end{aligned}
\label{eq:NF}
\end{equation}
where the first term on the right-hand side is the log-likelihood under the simple distribution, and $\mathbf{J}_{g_t^{-1}}(\vu_{t-1})$ is the Jacobian matrix of the inverse transformation $g_t$. To train the NF model and learn the source data distribution, the model parameters $\phi$ are typically optimized so to minimize the negative log-likelihood in Eq. \ref{eq:NF}. This results in the following loss function: %for the NF:
\begin{equation}
\gL_{NF} \, = \, -\log  p_{\mathrm{x}}(\vx)
\label{eq:NF_Loss}
\end{equation}
\mypar{Building the Normalizing Flow.} To build a bijective transformation function %neural network 
for the NF model, stacking a sequence of affine coupling layers~\cite{dinh2016density,kingma2018glow} has been demonstrated to be an efficient strategy. Because flows based on coupling layers are computationally symmetric, i.e., equally fast to evaluate or invert, they can overcome the %application and 
usability issues of asymmetric flows such as masked autoregressive flows, making them a popular choice. Let us consider $\vz \in \sR^D$ as the input to the coupling layer, which is split into a disjoint partition: ($\vz^A, \vz^B) \in \sR^d \times \sR^{D-d}$. %Furthermore, let $h_{\theta}(\cdot):\sR^d \rightarrow \sR^d$ denote a bijective function. 
The transformation function $g(\cdot):\sR^D \rightarrow \sR^D$ can then be defined as:
\begin{equation}
\vy^A =\vz^A, \quad \vy^B =\vz^B \odot \exp \left(s\left(\vz^A\right)\right)+t\left(\vz^A\right)
\label{eq:coupl}
\end{equation}
%where $\chi_{1}^{l}$ and $\chi_{2}^{l}$ correspond to disjoint sets of the input for the $l$-\textit{th} layer input, and $\chi_{1}^{l+1}$ and $\chi_{2}^{l+1}$ the corresponding disjoint output set. 
This setting offers simplicity for calculating the Jacobian determinant, which makes it possible to use complex neural networks as shift $s(\cdot)$ and scale $t(\cdot)$ networks. Note that the transformation in Eq. \ref{eq:coupl} is invertible and therefore allows for efficient Jacobian computation in Eq. \ref{eq:NF}. The work in \cite{dinh2016density} presented coupling flows on simpler tasks and datasets, e.g., CIFAR, which required less enriched representations. In contrast, the problem at hand requires pixel-to-pixel mappings on more challenging images. Thus, we replace the simple convolutional blocks in \cite{dinh2016density} with shallow U-shaped convolutional neural networks to find the shift and scale parameters of the affine transformation, as they capture more global context and provide higher representation power. Furthermore, as NFs are based on the change of variables rule, which is defined in continuous space, it is crucial to make the input continuous. Dequantization of the input can be achieved by adding a uniform noise $u\!\in\!U[0,1]$ to the discrete values. However, it might result in a hypercube representation of the images with sharp borders. These sharp borders are hard to model for a flow as it uses smooth transformations. Recently, a variational framework was proposed \cite{ho2019flow++} to extend dequantization to more sophisticated distributions, by replacing the uniform distribution with a learnable distribution.

\mypar{Constraining the source-distribution learning.} Optimizing the objective in Eq. \ref{eq:NF_Loss} with only source images might bias the model to focus on characteristics of subjects, such as age and gender, rather than on source-specific features like contrast and brightness. To overcome this issue, %we take inspiration from the outliers exposure technique in the task of anomaly detection \cite{hendrycks2018deep}. More concretely, we generate samples that differ from the source domain, potentially resulting in better network representations. Nevertheless, %the proposed method %to generate images far from the source domain 
%is fundamentally different from 
%the standard techniques to generate outliers cannot be applied in the task at hand. First, these techniques typically resort to real samples from other datasets to mimic an abnormal or out-of-distribution data point. Nevertheless, the proposed approach is domain-free, and cannot therefore access to any image from other domains than the source. And second, in the synthetic generation of outliers, Gaussian noise is typically added to original images, which simply adds noise but does not represent potential domain shifts. Our 
we propose a strategy that facilitates the learning of the source-domain distribution. This technique consists in randomly selecting $N'$ images from the original dataset $\gX_{\gS}$ and applying a series of augmentations $f_{aug}(\cdot)$ such that the resulting image has a dissimilarity to the original image (measured by mean squared distance) higher than a specified threshold. %$Th$. 
In particular, we employ contrast augmentation, brightness changes, multiplication, and random monotonically increasing mapping functions to augment these images. Then, the total learning objective of our model can be defined as:
\begin{equation}
\gL_T = 
%\left((
\underbrace{-\sum_{n=1}^{N-N'} \log  p_{\mathrm{x}}(\vx_n)}_{\textrm{Source distribution modeling}} - \underbrace{\sum_{n=1}^{N'} \min\left(\mathrm{c}, -\log  p_{\mathrm{x}}(f_{aug}(\vx_n))\right)}_{\textrm{Guiding term}}.
%\right)
\end{equation}
The first term is the learning objective in Eq. \ref{eq:NF_Loss} over the original source images, whereas the second one forces the NF model to decrease the likelihood on the augmented images, which facilitates the learning of domain-specific characteristics (e.g., contrast or brightness) instead of subject-related features (e.g., sex or age). Furthermore, we use a constant margin $\mathrm{c}$ in the second term to prevent the negative log-likelihood of an augmented sample from diverging to infinity.

\subsection{Achieving image harmonization}
\label{ssec:harmo}

\mypar{Harmonizer network.} A simple solution to perform image-to-image translation is to employ a harmonizer network $h_{\theta}(\cdot)$, such that MRIs from the target domain are translated to the source domain. This can be expressed as $p_{\mathbf{x}}(\mathbf{x})=p_{\mathbf{x}'}(h_{\theta}(\mathbf{x}'))$,
%The goal of our harmonizer network $h_{\theta}(\cdot)$ is to perform image-to-image translation of MRIs from the target to the source domain in such a way that:
%$$p_{\mathbf{x}}(\mathbf{x})=p_{\mathbf{x}'}(h_{\theta}(\mathbf{x}'))$$
where $\theta$ is the set of learnable parameters of the harmonizer network, and $\vx$ and $\vx'$ are images from the source and target domains, respectively. To train this model, we can simply use a standard reconstruction loss over images across different domains. However, we want the proposed method to follow a \textit{domain-free} paradigm, where target domains remain unknown at training time. Toward this goal, we train the harmonizer network to reconstruct the original source images from their augmented versions. As in the previous step, we augment the original images by using different types of contrast augmentation, brightness changes, multiplication, or random monotonically increasing mapping functions. Contrary to the first step, there is no constraint on the magnitude of the augmentations. The learning objective for the harmonizer network thus becomes:
\begin{equation}
{\theta^{init}} \, = \, \argmin _{\theta} \frac{1}{N} \sum_{n=1}^{N} \left\|(\vx_{n} - h_{\theta}\left(f_{aug}(\vx_{n})\right)\right\|^{2}
\label{eq:harmo}    
\end{equation}

We stress that the performed augmentations are not reliable representations of potential unseen target domains. Consequently, the direct application of the learned parameters $\theta^{init}$ for image-to-image mapping will result in suboptimal domain transformations. Nevertheless, they can serve as the initial model for the subsequent step. A simple UNet is considered for the harmonizer network, which learns two values. First, the last layer of the network ($\beta$) is employed as a bias value having the same dimension as the input image. Second, a scalar $\alpha$ from the middle layer of the network is used as a coefficient value. In this way, the output of the harmonizer can be defined as $h_{\theta}(\vx) = \alpha * \vx + \beta$.

%\subsection{Adapting harmonizer network with NF supervision.}

\mypar{Guiding the harmonizer network with the Normalizing Flow.}
%\textcolor{red}{The realistic assumption for harmonization is that the source domain is just accessible in the training time, and target domains are only accessible during the test time. in the training time, the harmonizer network is pre-trained with the source domain and the NF model learns the distribution of the source domain.}
The final step involves updating the harmonizer network so that images from the target domain are mapped into the source domain distribution. To achieve this, we propose to leverage the trained NF, which is stacked at the output of the harmonizer network. Note that the NF model has already learned the distribution of source data, and therefore its parameters remain frozen during the adaptation of the harmonizer. Thus, the learning objective of the adaptation stage consists in increasing the likelihood of the harmonizer outputs for images from the target domain, based on the NF model's density estimation. This loss function can be formally defined as follows:
\begin{equation}
\gL_{Adap} = 
-\sum_{m=1}^{M}\log  p_{\mathrm{x}}\big(g_{\phi}\big(h_{\theta}(\mathbf{x}_{m})\big)\big) 
\label{eq:adap_loss}    
\end{equation}

As stopping criterion for updating the harmonizer, we evaluate two possible alternatives. First, we measure the Shannon entropy of the predictions for the target task (e.g., segmentation or classification), stopping the adaptation when the entropy plateaus. We also consider the bits per dimension (\textit{bpd}), a scaled version of the \textit{negative log-likelihood} widely used for evaluating generative models:  $\textit{bpd}=- \log  p_{\mathrm{x}}(\vx) \cdot\left(\log 2 \cdot \prod_i \mathrm{\Omega}_i\right)^{-1}$ where $\mathrm{\Omega}_1$, ..., $\mathrm{\Omega}_T$, is the spatial dimension of the input images. More concretely, we can stop updating the harmonizer parameters when the reached \textit{bpd} value is the same as the one observed for the source domain using the NF model. In practice, this value can be obtained at training time using a validation set. 

%Another criterion that is used in this paper is updating the harmonizer's parameter until the entropy of the target task (segmentation, classification,..) is minimized. 

\section{Experiments}

\subsection{Experimental setting}
We evaluate the proposed method on the task of brain MRI segmentation across multiple sites. The reason behind this choice stems from the fact that %in addition to show how well target domains are harmonized, 
the segmentation performance is a reliable indicator of whether the structural information is well preserved during the mapping. 

\mypar{Datasets.} Four sites of the Autism Brain Imaging Data Exchange (ABIDE) \cite{di2014autism} dataset are employed: California Institute of Technology (CALTECH), Kennedy Krieger Institute (KKI), University of Pittsburgh School of Medicine (PITT) and NYU Langone Medical Center (NYU). The selection of these sites is based on their cross-site difference, as these datasets present the most distinct histogram from each other, which better highlights the impact of harmonization. These sites are denoted as $\mathcal{D}_1$, $\mathcal{D}_2$, $\mathcal{D}_3$, and $\mathcal{D}_4$, respectively. %Each site's scanner parameters and population age are shown in Table \ref{tab:Dataset-sites}.
From each site, we selected 20 T1-weighted MRIs from the healthy control population (19 from CALTECH), which were skull-stripped, motion-corrected, and quantized to 256 levels of intensity. %From these images, 60\% are used for training, 15\% for validation, and the remaining 25\% for testing, which are exploited in a 2D manner using the coronal plane. 
2D coronal slices of 60\% of these images are used for training, 15\% for validation, and the remaining 25\% for testing. Furthermore, the segmentation labels are obtained from FreeSurfer \cite{fischl2012freesurfer}, following other large-scale studies \cite{dolz20183d}, and grouped into 15 labels: background, cerebellum gray matter, cerebellum WM, cerebral GM, cerebral WM, thalamus, hippocampus, amygdala, ventricles, caudate, putamen, pallidum, ventral DC, CSF, and brainstem.

\mypar{Harmonization baselines.} The proposed approach is benchmarked against a set of relevant harmonization and image-to-image translation methods. We first consider a simple \textit{Baseline} applying the segmentation network directly on non-harmonized images, in order to assess the impact of each harmonization approach. Our comparison also includes: Histogram Matching \cite{nyul2000new}, aleatoric uncertainty estimation  (AUE) \cite{wang2019aleatoric}, Combat~\cite{pomponio2020harmonization}, BigAug~\cite{zhang2020generalizing} (which uses heavy augmentations for generalization of the segmentation networks), and two popular generative-based approaches, i.e., Cycle-GAN~\cite{modanwal2020mri} and Style-Transfer~\cite{liu2021style}.  

\mypar{Evaluation protocol.} To assess the  performance of our harmonization approach, we resort to a segmentation task as it requires the preservation of fine-grained structural details. First, a segmentation network $S_{\Phi}(\cdot)$ is trained on the images from the source domain, whose parameters remain frozen thereafter. The harmonized images from each method are then employed to evaluate segmentation performance, which is measured with the Dice Similarity Coefficient (DSC) and modified Hausdorff distance (HD). 
%after the harmonization of unseen domains.
To evaluate the robustness of tested methods, we repeat the experiments four times, each employing a different source and set of target domains. These different settings are denoted as $\gA: \gD_1\!\rightarrow\!\{\gD_2,\gD_3,\gD_4\}$; $\gB: \gD_2 \!\rightarrow\!\{\gD_1,\gD_3,\gD_4\}$; $\gC: \gD_3\!\rightarrow\!\{\gD_1,\gD_2,\gD_4\}$; $\gD: \gD_4\!\rightarrow\!\{\gD_1,\gD_2,\gD_3\}$.

\mypar{Implementation details.} \textbf{The Normalizing flow model} is trained for 1600 epochs using Adam optimizer with an initial learning rate of $1\times10^{-3}$, a weight decay of 0.5 every 200 epochs and a batch-size of 32. %images of size $224\times224$. 
We use a U-shaped network inside the coupling layers, which consists of four levels of different scales with a scaling factor of 2. Each level includes a modified version of the ELU activation function, i.e., concat(ELU($x$), ELU($-x$)), and a convolutional layer followed by a normalizing layer. %Also, the number of kernels of each level is 16, 32, 48, and 64, respectively. 
To construct the NF model, we first cascade four coupling layers with checkerboard masking to learn the noise distribution using variational dequantization. After applying four of the same coupling layers, features are squeezed as explained in \cite{dinh2016density} to have a lower spatial dimension and more channels. We then add four coupling layers using a channel-masking strategy, another feature squeezing function, and a final set of four coupling layers with channel-masking. The overall architecture of the flow model is shown in Fig. \ref{fig:normalizing flow architecture.}. The margin $\mathrm{c}$ used for guiding the flow is set empirically to 1.2. \textbf{The %U-shaped 
harmonizer} has five levels of different scales with a scaling factor of 2, each level including two layers of the modified ELU activation function followed by a convolutional layer. The number of kernels of each level is 16, 32, 48, 64, and 64, respectively. %, and both convolutions within the same level have the same number of kernels. %$\alpha$ coefficient is obtained using the Average pooling of the middle layer of the network followed by 64-kernel convolution, and $\beta$ value is the output of the last layer. 
The harmonizer is trained for 200 epochs using Adam optimizer with a learning rate starting at $1\times10^{-3}$, a weight decay of 0.5 every 30 epochs and a batch-size of 32. \textbf{The segmentation network} is trained for 200 epochs using Adam optimizer with an initial learning rate of $4\times10^{-3}$, a weight decay of 0.5 every 30 epochs and a batch-size of 32. All the models were implemented in PyTorch and were run on NVIDIA RTX A6000 GPU cards.

\subsection{Results}

\mypar{Comparison to state-of-the-art.} %To evaluate the proposed harmonization method, cross-site brain MRI segmentation performance has been obtained before and after the harmonization. As the segmentation networks are fixed, the segmentation results' improvement indicates the impact of the harmonization. These steps are repeated for other harmonization methods to compare the proposed methods with existing methods. Also, the results of test time augmentation \cite{wang2019aleatoric} (TTA) and using heavy augmentation for generalization of the segmentation networks (BigAug\cite{zhang2020generalizing}) have been obtained to investigate if they can replace the harmonization methods. 
Segmentation results obtained on the images harmonized by different methods are reported in Table \ref{tab:Results-comparison}. %\textcolor{red}{Before looking at the segmentation results, we would like to highlight that most existing methods make strong assumptions that might hamper their scalability and usability in real-life scenarios. First, some methods must access source images during the adaptation, thereby not being \textit{source-free} ($\gS\gF$). Second, labels associated with the downstream are required in some approaches, whereas we advocate for \textit{task-agnostic} ($\gT\gA$) methods. And last, most harmonization techniques need to access to the target domains during training, while ideally the potential target domains should remain unknown, which we refer to as \textit{unknown-domains} ($\gU\gD$). We relax all these assumptions by proposing a method that is \textit{source-free}, \textit{task-agnostic} and unaware of potential target domains during training.}
We can observe that the proposed approach consistently outperforms compared methods by a noticeable margin, across datasets and for both segmentation metrics. In particular, the average improvement gain is nearly 4\% in terms of DSC, and 0.7 $mm$ in terms of HD, compared to the second best performing method, CycleGAN. 
%the segmentation results' improvement indicates the impact of the harmonization. 

\begin{table}[t]
\caption{\label{tab:Results-comparison2} \textbf{Performance overview.} Main results for the compared methods across different settings ($\gA, \gB, \gC, \gD$). The best results are highlighted in bold.} %$\mathcal{A}: \mathcal{D}_1 \rightarrow \mathcal{D}_2,\mathcal{D}_3,\mathcal{D}_4$; $\mathcal{B}: \mathcal{D}_2 \rightarrow \mathcal{D}_1,\mathcal{D}_3,\mathcal{D}_4$;$\mathcal{C}: \mathcal{D}_3 \rightarrow \mathcal{D}_1,\mathcal{D}_2,\mathcal{D}_4$;$\mathcal{D}: \mathcal{D}_4 \rightarrow \mathcal{D}_1,\mathcal{D}_2,\mathcal{D}_3$. 
%$\mathcal{SF}$= Source-free, $\mathcal{TA}$=Task agnostic, $\mathcal{UD}$=Unknown domain.}
%\renewcommand{\arraystretch}{2}
\begin{center}
\scriptsize
%\makebox[\textwidth]
\setlength{\tabcolsep}{3pt}
{\begin{tabular}{l|ccc|c|c|c|c|c}
\toprule
%\textbf{Source}  &  \multicolumn{2}{c|}{$\mathcal{D}_1$}
%  &  \multicolumn{2}{c|}{$\mathcal{D}_2$}
%  &  \multicolumn{2}{c|}{$\mathcal{D}_3$}
%  &  \multicolumn{2}{c|}{$\mathcal{D}_4$}
%  &   \\ 
%  \midrule
%  \textbf{Target}  &  \multicolumn{2}{c|}{$\mathcal{D}_2,\mathcal{D}_3,\mathcal{D}_4$}
%  &  \multicolumn{2}{c|}{$\mathcal{D}_1,\mathcal{D}_3,\mathcal{D}_4$}
%  &  \multicolumn{2}{c|}{$\mathcal{D}_1,\mathcal{D}_2,\mathcal{D}_4$}
%  &  \multicolumn{2}{c|}{$\mathcal{D}_1,\mathcal{D}_2,\mathcal{D}_3$}
%  &  \multicolumn{2}{c}{Average} \\
% \midrule 
 & \multirow[b]{2}{*}{$\mathcal{SF}$} & 
 \multirow[b]{2}{*}{$\mathcal{TA}$} & 
 \multirow[b]{2}{*}{$\mathcal{UD}$} & $\mathcal{A}$ & $\mathcal{B}$ & $\mathcal{C}$ & $\mathcal{D}$ & Average\\
\cmidrule{5-9}
 & & & &  \multicolumn{5}{c}{DSC (\%)}\\
 \midrule 
Baseline  & -- & -- & -- & 54.6\ppm7.5 & 60.8\ppm4.6 & 62.9\ppm5.8 & 72.6\ppm4.5 & 62.7\ppm5.6 \\
AUE \cite{wang2019aleatoric} & \cmark & \xmark & \cmark & 54.7\ppm7.4 & 60.7\ppm4.7 & 62.6\ppm5.7 & 72.4\ppm4.5 & 62.6\ppm5.6 \\
Hist matching\cite{nyul2000new} & \cmark &  \cmark  & \cmark & 55.7\ppm8.6 & 58.1\ppm5.1 & 62.2\ppm4.8 & 69.5\ppm4.9 & 61.4\ppm5.9  \\
Combat \cite{pomponio2020harmonization} & \cmark &  \cmark & \cmark & 75.7\ppm9.2 & 79.9\ppm6.0 & 79.5\ppm8.1 & 79.9\ppm7.8 & 78.7\ppm7.8 \\
BigAug \cite{zhang2020generalizing} & \cmark & \xmark & \cmark & 54.2\ppm7.6 & 67.9\ppm3.6 & 61.5\ppm4.5 & 78.0\ppm3.7 & 65.4\ppm4.8 \\
Cycle-GAN \cite{modanwal2020mri}  & \xmark &  \cmark  & \xmark & 74.5\ppm3.0 & 78.8\ppm2.9 & 80.1\ppm2.2 & 83.1\ppm2.0 & 79.1\ppm2.5 \\
Style-transfer \cite{liu2021style}  & \cmark &  \cmark & \xmark & 56.9\ppm7.1 & 80.0\ppm1.7 & 67.8\ppm4.9 & 73.4\ppm4.0 & 69.5\ppm4.4    \\
Ours  &  \cmark &  \cmark  & \cmark & \textbf{80.8\ppm3.2}  & \textbf{82.3\ppm2.2} & \textbf{83.2\ppm3.3} & \textbf{85.2\ppm1.5}  & \textbf{82.9\ppm2.6} \\
\cmidrule{5-9}
 &\multicolumn{3}{c|}{} &  \multicolumn{5}{c}{HD (mm)}\\
\cmidrule{5-9}
Baseline  & -- & -- & -- & 18.20\ppm8.27 & 9.57\ppm3.23 & 9.07\ppm2.78 & 5.73\ppm1.81 & 10.64\ppm4.03  \\
AUE  \cite{wang2019aleatoric}& \cmark & \xmark & \cmark & 17.57\ppm8.18 & 9.67\ppm3.39 & 9.03\ppm2.87 & 5.57\ppm1.85 & 10.46\ppm4.08\\
Hist matching\cite{nyul2000new} & \cmark &  \cmark  & \cmark & 17.40\ppm8.10 & 10.47\ppm3.77 & 12.00\ppm4.56 & 6.73\ppm2.40 & 11.65\ppm4.71 \\
Combat \cite{pomponio2020harmonization} & \cmark &  \cmark & \cmark & 5.23\ppm3.87 & 3.67\ppm2.47 & 3.30\ppm1.80 & 3.17\ppm2.14 & 3.84\ppm2.57  \\
BigAug \cite{zhang2020generalizing} & \cmark & \xmark & \cmark & 19.53\ppm10.51 & 8.43\ppm3.40 & 18.87\ppm7.76 & 3.70\ppm1.07 & 12.63\ppm5.69 \\
Cycle-GAN  \cite{modanwal2020mri}& \xmark &  \cmark  & \xmark & 4.63\ppm2.89 & 3.63\ppm1.93 & 2.63\ppm0.62 & 2.30\ppm0.55 & 3.30\ppm1.50  \\
Style-transfer \cite{liu2021style} & \cmark &  \cmark & \xmark & 14.23\ppm7.20 & 2.93\ppm0.78 & 7.53\ppm2.38 & 4.27\ppm1.35 & 7.24\ppm2.92  \\
Ours  &  \cmark &  \cmark  & \cmark & \textbf{3.10\ppm1.63} & \textbf{2.77\ppm0.87} & \textbf{2.37\ppm0.77} & \textbf{2.30\ppm0.50} & \textbf{2.63\ppm0.94} \\
\bottomrule
\end{tabular}}
\end{center}

\label{tab:Results-comparison}
\end{table}

\mypar{Impact of normalizing flows.} 
This section assesses the impact of each component of the proposed method. In particular, we evaluate the segmentation performance when images are: \textit{i)} not normalized, \textit{ii)} normalized with the pre-trained harmonizer $\theta^{init}$, or \textit{iii)} normalized with the proposed harmonizing flow. The results from this ablation study (Table \ref{tab:Results-impact-of-HF-part}) empirically motivate the proposed NF model as a powerful mechanism to guide the harmonizer network. First, the strategy proposed to pre-train the harmonizer brings a substantial improvement over non-harmonized images, yet it is very simple and does not require access to images from the target domain. Secondly, driving the adaptation of the harmonizer with the proposed NF further improves the segmentation results by a large margin, demonstrating the benefits of our model.

%, in particular, the harmonizing network and NF model. This was accomplished by first comparing the segmentation results obtained from baseline images with those obtained from the pre-trained harmonizer network. The performance improvement is indicative of the effects of pre-trained harmonization. In the next step, the harmonizer network is adapted to the target domain with the NF model supervision. As is shown in Table \ref{tab:Results-impact-of-HF-part}, the pre-trained harmonizer network gains 12.4\% DSC improvement over the baseline on average, and the NF model gains an additional 7.8\%.

\begin{table}[h]
\caption{\label{tab:Results-impact-of-HF-part} Ablation study on the different components in terms of DSC.}
\begin{center}
\footnotesize
\makebox[\textwidth]{
\setlength{\tabcolsep}{3pt}
{\begin{tabular}{l|c|c|c|c|c}
\toprule
 & $\mathcal{A}$ & $\mathcal{B}$ & {$\mathcal{C}$} &  $\mathcal{D}$ & Average\\
 \midrule
Without harmonization & 54.6\ppm7.5 & 60.8\ppm4.6 & 62.9\ppm5.8 & 72.6\ppm4.5 & 62.7\ppm5.6 \\

Pre-trained harmonizer & 71.9\ppm5.1 & 77.0\ppm3.0 & 76.0\ppm5.2  & 75.3\ppm4.5 & 75.1\ppm4.5 \\

Adapting using NF & 80.8\ppm3.2 & 82.3\ppm2.2 & 83.2\ppm3.3 & 85.2\ppm1.5  & 82.9\ppm2.6 \\

% \cmidrule{2-6}
% &  \multicolumn{5}{c}{HD95}\\
% \cmidrule{2-6}

% Without harmonization & 18.2\ppm8.3 & 9.6\ppm3.2 & 9.1\ppm2.8 & 5.7\ppm1.8 & 10.6\ppm4.0 \\

% Pre-trained harmonizer & 6.5\ppm3.0 & 3.8\ppm1.5 & 4.5\ppm1.7 & 4.4\ppm1.3 &  4.8\ppm1.8 \\

% Adapting using NF & 3.1\ppm1.6 & 2.8\ppm0.9 & 2.4\ppm0.8 &  2.3\ppm0.5 &  2.6\ppm0.9 \\
\bottomrule
\end{tabular}}}
\end{center}
\end{table}

\mypar{Adaptation stopping criterion.} In this section, we address the important question of when to stop the adaptation. %, as the harmonizer updates are done in an unsupervised manner. 
The first alternative is to stop it when the Shannon entropy of the segmentation predictions reaches its minimum point. As this objective does not require any labeled data, it gives a valid stopping point for adapting the harmonizer. As a second criterion, we stop adapting when the output \textit{bpd} of the NF model reaches the observed source domain \textit{bpd}. As opposed to entropy, this criterion is not task-dependent and is suitable for unsupervised tasks %(e.g., unsupervised anomaly detection)
or tasks where entropy is not applicable. Last, we resort to the segmentation performance, and stop the adaptation when it reaches the best DSC score, which we define as \textit{Oracle}. Note that this criterion is unrealistic, and its purpose is just to demonstrate how a good stopping criterion can improve harmonization. As shown in Table \ref{tab:Results-compare-stopping-criterion}, although minimum entropy is a better criterion compared to \textit{bpd}, both achieve comparable performances. In addition, both stopping criteria are a suitable choice, as their results are very close to the \textit{Oracle}.

\begin{table}[h]
\caption{\label{tab:Results-compare-stopping-criterion} Impact of the adaptation stopping criterion (in terms of DSC).}
\begin{center}
\footnotesize
\makebox[\textwidth]{
\setlength{\tabcolsep}{3pt}
{\begin{tabular}{l|c|c|c|c|c}
\toprule
 & $\mathcal{A}$ & $\mathcal{B}$ & {$\mathcal{C}$} &  $\mathcal{D}$ & Average\\
 \midrule
Minimum Entropy & 80.8\ppm3.2 & 82.3\ppm2.2 & 83.2\ppm3.3 & 85.2\ppm1.5 & 82.9\ppm2.6 \\
Source BPD &  80.5\ppm3.1 & 82.6\ppm2.3 & 82.7\ppm3.6  & 84.8\ppm1.6 & 82.6\ppm2.6  \\
Oracle (best epoch) &  81.0\ppm3.0 & 82.7\ppm2.3 &  84.0\ppm2.9 & 85.2\ppm1.5 & 83.2\ppm2.4 \\

% \cmidrule{2-6}
% &  \multicolumn{5}{c}{HD95}\\
% \cmidrule{2-6}

% Minimum Entropy & 3.1\ppm1.6 & 2.8\ppm0.9 & 2.4\ppm0.8 & 2.3\ppm0.5 & 2.6\ppm0.9 \\
% Source BPD & 3.6\ppm1.8  & 3.1\ppm1.6 & 2.4\ppm0.8  & 2.4\ppm0.4 & 2.9\ppm1.2 \\
% Oracle (best epoch) &  3.0\ppm1.6  & 3.2\ppm1.3 &  2.2\ppm0.6 &  2.3\ppm0.5  & 2.7\ppm1.0  \\
\bottomrule
\end{tabular}}}
\end{center}
\end{table}

\begin{figure}[t]
\begin{center}    
    {\includegraphics[width=\textwidth]{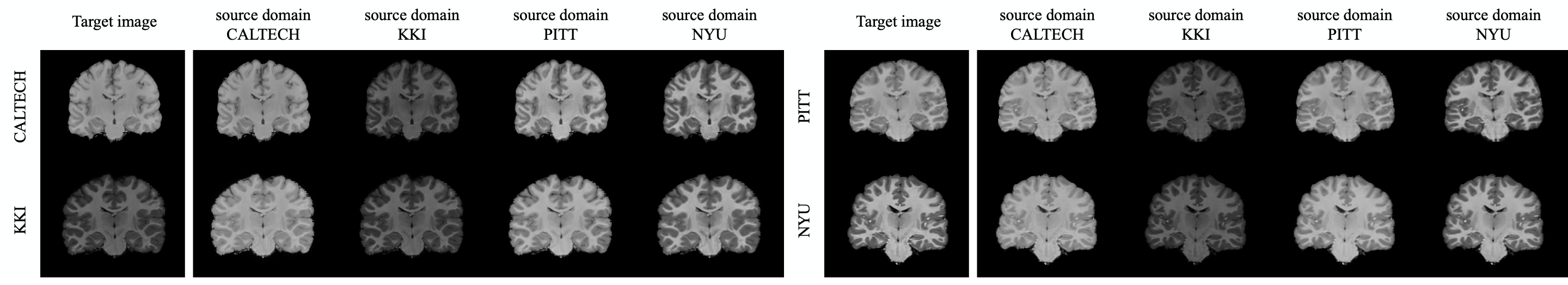}}
    \caption{Examples of harmonized images produced by the proposed method.}
    \label{fig:results}
\end{center}
\end{figure}

\mypar{Qualitative results.} Figure \ref{fig:results} depicts several examples of harmonized images produced by the proposed approach. These results illustrate that, regardless of the target domain, our method produces reliable image-to-image mappings to the source distribution. 
  
\begin{table}[h]
\caption{\label{tab:Results-comparison-N4} Quantitative results on bias corrected images (in terms of DSC).}
\begin{center}
\footnotesize
\makebox[\textwidth]{
\setlength{\tabcolsep}{3pt}
{\begin{tabular}{l|c|c|c|c|c}
\toprule
 & $\mathcal{A}$ & $\mathcal{B}$ & {$\mathcal{C}$} &  $\mathcal{D}$ & Average\\
 \midrule
Baseline  &  71.9\ppm3.7 &  78.2\ppm3.6 &  82.9\ppm1.7 &  82.6\ppm2.9 &  78.9\ppm3.0 \\
 AUE \cite{wang2019aleatoric}  &  71.4\ppm3.7 &  78.3\ppm3.6 &  82.7\ppm1.7 &  82.7\ppm2.8 &  78.8\ppm2.9 \\
 Hist matching \cite{nyul2000new} &  78.7\ppm1.9 &  80.6\ppm2.4 &  83.1\ppm2.0 &  81.3\ppm2.6 &  80.9\ppm2.2 \\
 Combat \cite{pomponio2020harmonization} &  77.4\ppm1.6 &  80.2\ppm1.3 &  79.8\ppm2.0 &  78.6\ppm2.0 &  79.0\ppm1.7  \\
 BigAug \cite{zhang2020generalizing}  &  81.7\ppm2.1 &  82.4\ppm1.5 &  85.2\ppm0.9 &  84.6\ppm1.3 &  83.5\ppm1.4  \\
 Cycle-GAN \cite{modanwal2020mri}  &   78.3\ppm2.0 &  79.9\ppm1.0 &  82.9\ppm1.3 &  83.4\ppm1.0 &  81.1\ppm1.3  \\
 Style-transfer \cite{liu2021style} &   74.1\ppm3.0 &  80.0\ppm1.3 &  80.6\ppm1.5 &  81.1\ppm1.5 &  79.0\ppm1.8  \\
 Ours  & \textbf{83.0\ppm1.8} &  	\textbf{84.4\ppm1.5} & \textbf{85.4\ppm1.2} &  	\textbf{85.6\ppm1.3} & \textbf{84.6\ppm1.5} \\

% \cmidrule{2-6}
% &  \multicolumn{5}{c}{HD95}\\
% \cmidrule{2-6}

% Baseline  &  7.13\ppm1.85 &  4.80\ppm1.76 &  4.27\ppm1.37 &  3.97\ppm0.67 &  5.04\ppm1.41  \\
%  TTA \cite{wang2019aleatoric}  &  6.77\ppm1.94 &  4.27\ppm1.18 &  4.20\ppm1.31 &  3.83\ppm0.68 &  4.77\ppm1.28  \\
%  Hist matching \cite{nyul2000new} &  5.00\ppm1.57 &  4.43\ppm1.31 &  4.27\ppm1.71 &  4.10\ppm0.72 &  4.45\ppm1.33   \\
%  Combat \cite{pomponio2020harmonization} &  4.50\ppm1.70 &  2.70\ppm0.32 &  2.87\ppm0.39 &  3.07\ppm0.58 &  3.28\ppm0.75  \\
%  BigAug \cite{zhang2020generalizing}  &  \textbf{2.60\ppm0.31} &  2.87\ppm0.46 &  2.47\ppm0.77 &  2.63\ppm0.34 &  2.64\ppm0.47  \\
%  Cycle-GAN \cite{modanwal2020mri} &  3.20\ppm0.74 &  2.53\ppm0.15 &  2.30\ppm0.21 &  2.17\ppm0.13 &  2.55\ppm0.31 \\
%  Style-transfer \cite{liu2021style} &  5.57\ppm1.47 &  2.73\ppm0.64 &  3.83\ppm1.53 &  2.97\ppm0.64 &  3.77\ppm1.07  \\
%  Ours  & 3.00\ppm0.97 & \textbf{2.27\ppm0.60} & \textbf{2.27\ppm0.73} & \textbf{1.90\ppm0.17} & \textbf{2.36\ppm0.62} \\
\bottomrule
\end{tabular}}}
\end{center}
\end{table}

\mypar{Results when N4 bias correction is applied.} 
In previous sections, we used the original MRIs of the ABIDE dataset without bias correction to evaluate the proposed harmonization method on more challenging scenarios, where pre-processing steps to enhance the images might not be applicable. Compared to bias-corrected MRIs, original MRIs have arguably more complex distributions, which makes it more difficult for harmonization methods to map MRIs from a target domain to the source one. To demonstrate that our method also achieves satisfactory performance when the initial domain shifts are reduced, we repeated the previous steps with N4-biased corrected MRIs. These results, shown in Table \ref{tab:Results-comparison-N4}, also showcase the advantage of our method in this different setting. For conciseness, we report here the average results for the HD metric (in $mm$) across different methods: Baseline ($5.04\pm1.41$), Hist matching ($4.45\pm1.33$), Combat ($3.28\pm0.75$), BigAUG ($2.64\pm0.47$), Cycle-GAN ($2.55\pm0.31$), Style-transfer ($3.77\pm1.07$) and ours ($2.36\pm0.62$).

\mypar{Experiments on Test-Time Adaptation (TTA).} Our model can also be employed in a TTA scenario, where the model needs to be updated at inference for a given image, or set of images. To motivate this assumption, we compare the performance of our approach to the popular TENT model~\cite{wang2020tent}. %, which explicitly minimizes the entropy of the test image predictions. 
Adapting the segmentation network $\gS_{\Phi}(\cdot)$ with TENT yields $65.1\pm5.0$ of DSC, which represents a considerable gap compared to our model, i.e., $82.9\pm2.6$. Note that there exist other TTA methods for segmentation in the medical field, e.g., \cite{karani2021test}, however, they require segmentation masks for the adaptation. 

%%% JOSE: To CHECK %%%
\section{Conclusion}
In this paper, we proposed a novel harmonization method which leverages Normalizing Flows to guide the adaptation of a harmonizer network. Our approach is source-free, task-agnostic, and works with unseen domains. These characteristics make our model applicable in real-life problems where %because of privacy, 
the source domain might not accessible during adaptation, target domains are unknown at training time, and harmonization is not dependent on a specified target task. %and can be applied as a pre-processing step. 
%Furthermore, an important advantage of our method over the existing approaches is that it only requires images from the source domain, and one modality, at training time.
%this method only needs one source domain and modality in the training time, which is an important advantage over the existing approaches. 
%In particular, our approach leverages Normalizing Flows to guide the adaptation of a harmonizer network. 
The proposed method achieves state-of-the-art harmonization performance based on the segmentation task, yet relaxes the strong assumptions made by existing harmonization strategies.
%While the proposed method offers the aforementioned characteristics, results show that it also achieves decent performance when evaluated using cross-site MR brain segmentation, and the results are superior to the other existing approaches. Also, as is demonstrated in the results, both proposed stopping criteria seem to be suitable, and the normalizing flow model have a great impact on adapting the harmonizer network. 
%Taking into account that our approach is \textit{source-free}, does not require labels of the downstream task, and can generalize to multiple domains without accessing them during training, the results reported in the experimental section empirically position our model as a powerful alternative for MRI multi-site harmonization. 
Thus, we believe that our model is a powerful alternative for MRI multi-site harmonization.

%%%%%%%%%%%%%%%%%%%%%%%%%%%%%%%%%%%%%%%%%%%%%%%%%%%%%%%%%%%%

 \bibliographystyle{splncs04}
 \bibliography{mybibliography}

\end{document}